\documentclass[10pt, conference, compsocconf]{IEEEtran}
\ifCLASSINFOpdf
\else
\fi
\usepackage[utf8]{inputenc}
\usepackage{gensymb}
\usepackage{graphicx}
\graphicspath{ {./} }
\usepackage{multirow}

\begin{document}
%
\title{Face Detection in Camera Captured Images of Identity Documents under Challenging Conditions}



\author{Souhail Bakkali \quad Zuheng Ming  \quad Muhammad Muzzamil Luqman \quad Jean-Christophe Burie\\
    L3i, La Rochelle University, France\\
    \{souhail.bakkali, zuheng.ming, jcburie, mluqma01\}@univ-lr.fr}
\maketitle

\begin{abstract}
Benefiting from the advance of deep convolutional neural network approaches (CNNs), many face detection algorithms have achieved state-of-the-art performance in terms of accuracy and very high speed in unconstrained applications. However, due to the lack of public datasets and due to the variation of the orientation of face images, the complex background and lighting, defocus and the varying illumination of camera captured images, face detection on identity documents under unconstrained environments has not been sufficiently studied. To address this problem more efficiently, we survey three state-of-the-art face detection methods based on general images, i.e. Cascade-CNN, MTCNN and PCN, for face detection in camera captured images of identity documents, given different image quality assessments. For that, The MIDV-500 dataset, which is the largest and most challenging dataset for identity documents, is used to evaluate the three methods. The evaluation results show the performance and the limitations of the current methods for face detection on identity documents under the wild complex environments. These results show that the face detection task in camera captured images of identity documents is challenging, providing a space to improve in the future works.

\end{abstract}

\begin{IEEEkeywords}
Face detection, Identity Document, Deep CNNs

\end{IEEEkeywords}

\IEEEpeerreviewmaketitle

\section{\textbf{Introduction}}
\begin{figure}[t]
\begin{center}
   \includegraphics[width=1.0\linewidth]{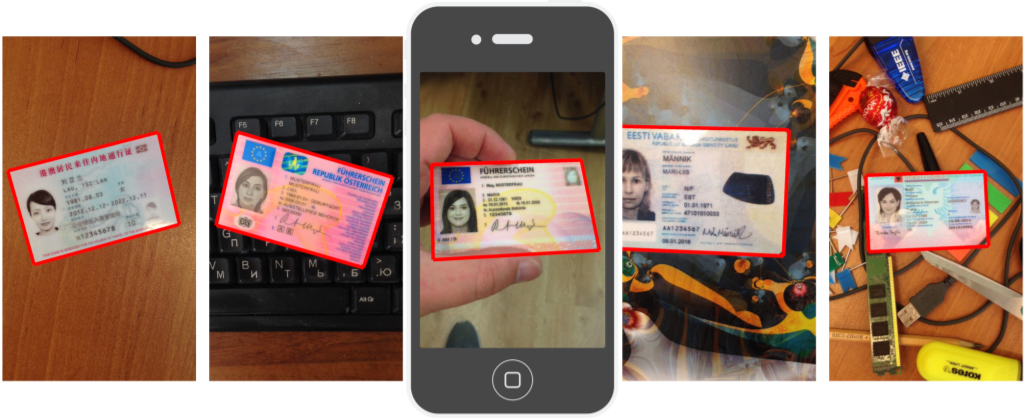}
\end{center}
   \caption{Frame examples from MIDV-500 dataset. Different conditions are presented such as different orientation, complex background from different scenarios and the variation of the illumination.}
\label{fig:im1}
\end{figure}

A lot of researches has been dedicated and reserved to identity document analysis and recognition on mobile devices, willing to facilitate the data input process \cite{de2016money}. A practical and common approach to this problem involves detecting and comparing individual's live face to the face image found in his/her identity document. However, the face detection and recognition on the identity documents poses numerous challenges that are different from general face detection and recognition on the normal visual images. 
The main challenges of face detection in images on identity documents are mainly due to the challenging conditions in which camera captured images on identity documents have been taken, such as occlusion, defocus, complex background, the variation of the orientation of identity documents and the varying illumination conditions caused by the reflect. ~\figurename\ref{fig:im1} shows the challenging conditions of camera captured images on identity documents from the MIDV-500 dataset \cite{arlazarov2018midv}. 
Besides, collecting information from identity documents (such as faces, signatures, texts..) is still a challenge to solve due to the strict privacy laws and regulations. Since identity documents contain sensitive personal data, people are not willing to risk the leak of the personal information that may lead to some complications. 
Thus, it becomes difficult to evaluate and compare various identity document analysis methods to each other, since they have been tested on private industry-oriented and locked down data \cite{skoryukina2015real, usilin2010visual}, which are collected from customers and are not available for the public in the light of security and market advantage. Therefore, the lack of public datasets for images in identity documents also hinder the research on this sensitive field.
Thanks to the availability of large face datasets and the progress in deep neural network architectures \cite{hu2017finding, najibi2017ssh, yang2015facial}, face detection and recognition in general face images has made tremendous strides in the past decade. Recently, there have been a necessity to build a model to accurately differentiate faces from the backgrounds in challenging conditions, while keeping real-time performance. A deep cascaded multi-task architecture built on deep convolutional neural networks (CNNs) employs a coarse-to-fine strategy for face detection, by adopting different stages of CNNs in order to predict the face progressively~\cite{shi2018real,zhang2016joint, li2015convolutional}. Benefiting from the simplicity of the global networks and the progressive increase of depth of the CNNs, the deep cascaded based CNNs for face detection can achieve the state-of-the-art performance in terms of accuracy and speed. Specifically, Cascade-CNN~\cite{li2015convolutional} firstly proposed the deep Cascade-CNNs architecture for face detection task. The MTCNN~\cite{zhang2016joint}, based on the deep cascaded multi-task framework, achieved also the state-of-the-art performance and attained very high speed for face detection and alignment. Furthermore, PCN~\cite{shi2018real} which is also based on the deep cascaded CNNs, has tackled the detection of rotating faces in a coarse-to-fine manner. It can accurately detect faces with arbitrary rotation-in-plane angles, which are common to the camera captured images on identity documents. In this work, these three state-of-the-art methods Cascade-CNN, MTCNN and PCN, that are the most widely used for face detection tasks with different advantages, are adopted to detect faces on identity document images. The recent published MIDV-500 dataset, which is a Mobile Identity Document Video dataset consisting of 500 video clips for 50 different identity with 17 types  of ID cards, is used to evaluate the three different face detection frameworks.


In summary, our main contributions of this paper are:
\begin{itemize}
	\item Evaluating the three state-of-the-art methods Cascade-CNN, MTCNN and PCN for face detection in camera captured images on identity documents under challenging environments. 
    \item The bounding box coordinates of face regions in arbitrary frames in the MIDV-500 dataset's videos have been manually annotated for the evaluation.
    \item We have demonstrated that, under the challenging conditions, with various orientation of the document, complex background and the varying illumination, the MTCNN model shows its superiority to the other methods evaluated on the MIDV-500 dataset.
\end{itemize}

The rest of paper is organized as follows: in Section II, we briefly review the existing studies in the area of identity document analysis and recognition; Section III describes the frameworks of the evaluated methods; in Section IV, we present the Mobile Identity Document Video dataset (MIDV-500) \cite{arlazarov2018midv} and the experimental evaluation results; the final Section V draw a conclusion and presents the future work. 

\section{\textbf{Related Work}}
Lately, numerous researches have been devoted to document analysis and recognition, analyzing and identifying different identity fields: document numbers, document holder name components, machine readable zone, signatures and face detection using state-of-the-art visual recognition approaches \cite{awal2017complex,simon2015fine,de2015use}. As for the related work provided on \cite{arlazarov2018midv}, a text field OCR study was proposed to perform text field extraction, recognition and identity document data extraction from video clips. 

The face detection problem has been an important task in computer vision systems that aims to extract information from face images, with respect to the head position and occlusion, while detecting faces quickly and accurately on input images. Previous face detection systems were mostly based on hand-crafted features extracted by the feature engineering for the classification methods. A number of variant methods and local descriptors have been proposed for face detection and recognition task, such as, rapid object detection using a boosted cascade of simple features~\cite{viola2001rapid}, LBP~\cite{cinbis2011unsupervised}, HOG~\cite{wolf2011face}, Gabor-LBP~\cite{sivic2005person}, SIFT \cite{lu2015surpassing}. Recent researches in this area focus more on the uncontrolled face detection problem, where various poses, complex lighting, occlusion, exaggerated expressions, full rotation-in-plane angle face images can lead to large visual variations, and to significant divergence in face appearances, while detecting faces in full rotation-in-plane angles can greatly advance the performance of both face alignment and face detection-recognition.

Nevertheless, the first study on ID document face photo is attributed to Starovoitov et al. \cite{starovoitov2002three,starovoitov2000matching}, assuming that all face images are frontal faces without large expression variations as most methods do. The algorithm is similar to a general constrained face matcher, except that it is developed for a document photo dataset. Since the popularity of deep neural networks could partially be attributed to a special property that the low-level image features are transferable, i.e. they are not limited to a particular task, but applicable to many image analysis tasks. Given this property, one can first train a domain specific neural network by transfer learning on a relatively small dataset, as proposed by Y. Shi and Anil K. Jain \cite{shi2018docface}. A recent face detection study was performed in \cite{arlazarov2018midv} using open source libraries \cite{king2009dlib, bradski2000opencv} with default frontal face detectors. It was conducted using original frames, and using projectively restored document images based on ground truth document coordinates. The ground truth coordinates were projectively transformed from the template coordinates to frame coordinates (according to the ground truth document boundaries). The same process was done for face detection results in the cropped document detection mode. \\

\section{\textbf{Face Detection Frameworks}} 
The frameworks have been chosen for their performance in terms of high accuracy and speed, their low time cost in a fast run-time on the Multi-oriented FDDB \cite{jain2010fddb} and WiderFace \cite{yang2016wider} datasets, and also for their specific architecture based on deep cascaded multi-task convolutional neural networks (CNNs). 
High accuracy is achieved with a deep neural network; having three networks - each with multiple layers - allow for higher precision, as each network can fine-tune the results of the previous one. This technique is so called Hard Sample Mining. As for the three methods, training only the first stage is not perfect, because it would recognize some images with no faces in it as positive samples (window candidates containing at least one face). These images are known as false positives, so we include them into The second stage, since its role is to refine bounding box edges and reduce false positives. This can help the second stage targets the first network's weaknesses and improve accuracy. Similarly, it is applied to the third stage as well. 

Since different tasks are employed in each CNNs, different types of training images are used in the learning process, such as faces, non-faces and partially aligned faces, and instead of defining only one loss function for all tasks, one loss function each is defined, they are switched back and forth with different ratios to balance different loss functions, according to the task importance.

As for the training data used to train the models, three kinds of images are employed: positive samples, negative samples and suspected samples. Positive samples are those windows with IoU ratio over 0.7/0.65 to any ground truth faces; negative regions are those windows with IoU ratio smaller than 0.3 to a ground truth face, and suspected/part samples are those windows with IoU between 0.4 and 0.7/0.65. Positive and Negative samples are used for the classification task of faces/non-faces. Positive and suspected/part samples contribute to the training of bounding box regression and calibration. 

\subsection{\textbf{Cascade-CNN face detector}} 

The Cascade-CNN face detector~\cite{shi2018real}, compared to other face detection systems, that learns the classifier by relying on hand-craft features, evaluates the input given image to reject non-face regions and distinguishes faces from high covered regions at higher resolution. After that, a calibration network is applied to process the remaining detection windows to adjust its size and calibrate its location to approach a potential face. The proposed cascade face detector is a three-stage deep convolutional network. After each stage, Non-Maximum Suppression is introduced to eliminate highly overlapped detection windows to make a more accurate detection window at the correct scale, with an Intersection-over-Union ratio exceeding a set ratio. The remaining calibrated bounding boxes are considered as the outputs of the model. 

The experiments on the the challenging Face Detection Dataset and Benchmark FDDB show that the Cascade-CNN detector outperform the state-of-the-art methods in the continuous score evaluation, and is comparable to the state-of-the-art methods on the Annotated Faces in the Wild AFW \cite{zhu2012face}.

\subsection{\textbf{Joint Face Detection and Alignment}}
The proposed MTCNN framework ~\cite{zhang2016joint} uses multi-task cascaded convolutional networks with three stages, to predict face and landmark location in a coarse-to-fine manner. The cascaded convolutional network aims to reduce the extra computational expense of the cascade face detector proposed by Viola and Jones \cite{viola2001rapid}, it utilizes Haar features with boosted cascade framework to evaluate frontal faces, but the framework is relatively weak towards uncontrolled applications where faces are in varied poses and complex unexpected lighting. The need of convolutional networks was necessary to achieve remarkable performance in a variety of computer vision tasks. The contribution of this method is to combine face alignment and face detection for real time performance.

The overall framework of this approach consists of obtaining different scaled images by passing in the sliding window and image pyramid, each candidate window goes through the detector stage by stage. In each stage, the detector rejects faces with low confidence level, regresses the bounding boxes of remaining faces. For the remaining face candidates, they are updated to the new bounding boxes that are regressed. After each stage, non-maximum-suppression is performed on every box to merge those highly overlapped face candidates. The output of the last stage will sort bounding boxes of the remaining calibrated face candidates with facial landmarks' positions, having only one bounding box for every face in the image. 

The performance of the model was evaluated on FDDB and WIDER FACE datasets, outperforming the state-of-the-art methods by a large margin in both benchmarks.

\subsection{\textbf{Progressive Calibration Networks (PCN)}}
The proposed progressive calibration network (PCN) ~\cite{li2015convolutional} is a real-time and accurate face detector, which aims to progressively calibrate the rotation-in-plane angle of each face candidate to upright in a three-stage multi-task deep convolutional network. More specifically, the calibration process to precise the rotation-in-plane angle is divided into several progressive steps and only predicts the coarse orientation in each stage.
Given an input image, a sliding window and image pyramid principle are applied to obtain and detect all different sized faces within the image. Each face will be passed through the detector stage by stage. After passing in the image, the detector gathers face candidates and their bounding box coordinates, parse the stage output to get a list of confidence levels for each bounding box, then rejects most candidates with low face confidence, i.e. (Boxes that the network is not quite sure contains a face). Simultaneously, the estimated bounding boxes of remaining face candidates are regressed and calibrated according to the predicted coarse rotation-in-plane angles. After each stage, non-maximum-suppression is conducted to eliminate redundant boxes. The calibration based on the coarse rotation-in-plane prediction brings almost no additional time cost, leading to accurate and fast calibration.

The experiments on the multi-oriented FDDB and Rotation WIDER FACE datasets show that PCN performs way better than the baseline Cascade, with almost no extra time cost benefited from the efficient calibration process.

\section{\textbf{Experimental Baselines}}
In the following part, we decribe at first the MIDV-500 dataset. After that, we present the evaluation results on the dataset to demonstrate the effectiveness and the limits of the methods described above.  

\subsection{\textbf{Dataset Structure}}
 The MIDV-500 is a Mobile Identity Document Video dataset consisting of 500 video clips for 50 different identity document types with ground truth, including 17 types of ID cards, 14 types of passports, 13 types of driving licences and 6 other identity documents of various countries. The video clips were recorded in 5 different conditions: \\
 
\textbf{CA}, \textbf{CS}: "Clutter" - The document lays on table with many objects in the background.\\
\textbf{HA}, \textbf{HS}: "Hand" - The document is held on hand.\\
\textbf{KA}, \textbf{KS}: "Keyboard" - The document lays on keyboards.\\
\textbf{PA}, \textbf{PS}: "Partial" - The document is partially or completely hidden off-screen.\\
\textbf{TA}, \textbf{TS}: "Table" - The document is presented on a table.\\

The dataset has about (50 documents) x (5 conditions) x (2 devices), each video was split into 10 frames per second while the duration of each video is 3 seconds, which makes it (30 images) x (2 devices) for every condition. In total there are 15 000 annotated frame images in the whole dataset. 

For each extracted video frame, bounding box annotation for each face was performed manually. In total, there are 48 ID-faces out of 50 identity documents.
The MIDV-500 dataset allows to perform studies of object detection and information extraction algorithms, measures their accuracy and their robustness against various distortions. For our study, we regard the face detection experiment to evaluate the state-of-the-art frameworks described above. For that, we consider only a part of the dataset taking into account the presence of a face on the document. Since each condition contains 30 frames per 3 seconds, most frames remain the same. So, we have taken only 2 frames per condition to evaluate the robustness of the three models. This way, 1000 frames (50 document) x (5 conditions) x (2 devices) x (2 frames) out of 15 000 were filtered for the experiment. 

\subsection{\textbf{Results on MIDV-500 Dataset}}

\begin{figure}[t]
\begin{center}
   \includegraphics[width=0.9\linewidth]{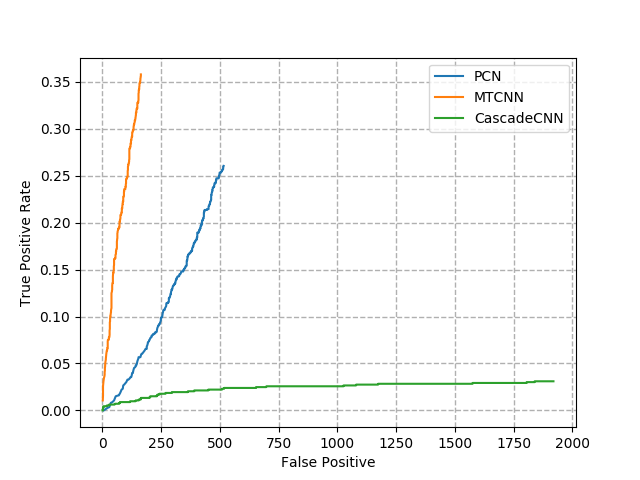}
\end{center}
   \caption{ROC curves of three face detectors PCN, MTCNN and Cascade-CNN on MIDV-500. The horizontal axis on the ROC curve is the number of “False positives over the whole dataset” and the vertical axis on the ROC curve is the "True positive rate", i.e. recall rate. Usually, 20 to 200 FP is a sensible operating range in the existing works~\cite{opitz2016grid}.}
\label{fig:ROC}
\end{figure}

\begin{figure*}[t]
\begin{center}
   \includegraphics[width=1.0\linewidth]{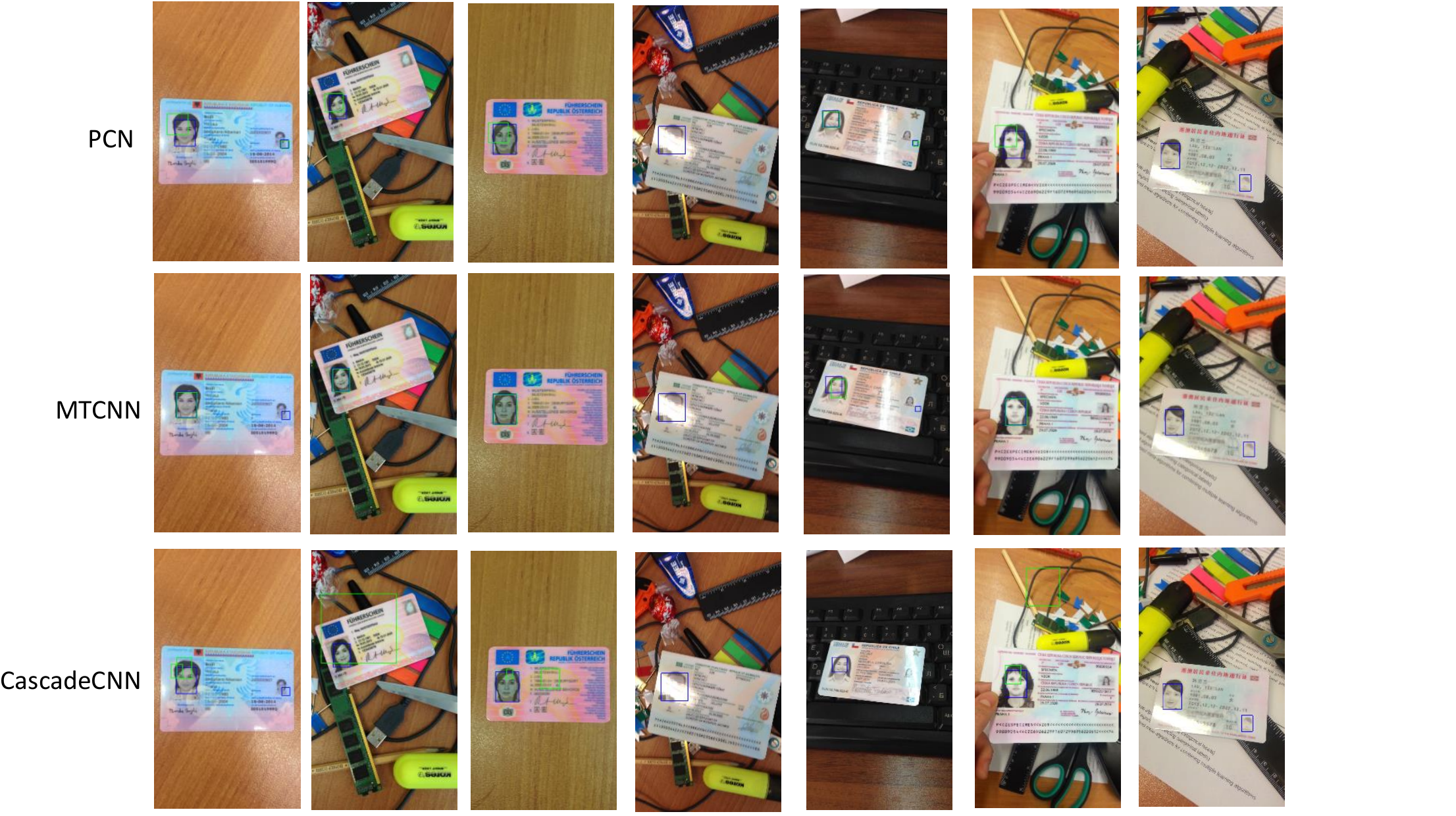}
\end{center}
   \caption{Some examples of the detected faces by PCN, MTCNN and CascadeCNN respectively. The rectangle (i.e. bounding box) in blue is the ground truth and the rectangle in green is the detected bounding box.}
\label{fig:detectionexamples}
\end{figure*}

Face detection was performed using Cascade-CNN, MTCNN and PCN. The detection was conducted using original frames, where faces presented on the document images were annotated manually. We haven't conducted face detection on projective restoration of document images based on ground truth document coordinates, since our objective is to perform the robustness of the three models on various poses and different face appearances in different angles. Following the protocol proposed by~\cite{jain2010fddb} of the evaluation of the face detector, we evaluate the three face detectors mentioned above in terms of ROC curves on MIDV-500 dataset shown in ~\figurename~\ref{fig:ROC}. In the ROC curve, the 20 to 200 FP is usually a sensible operating range in the existing works~\cite{opitz2016grid}. We can see that the MTCNN (red curve) is the best face detector on MIDV-500 dataset, then the PCN (on blue curve) and the Cascade-CNN (green curve) performs worst. Even if the PCN face detector is designed for the rotation-invariant face detection, the PCN does not show its superiority on MIDV-500 dataset. As well as the PCN, the Cascade-CNN can also detect the rotating face images by using four small CNNs to deal with four directions of the rotation, i.e. up, down, right, left. However for each direction, the CNN used for the detection is small and the performance is inferior than the other detectors. Overall, MTCNN shows a good generalization ability on the new challenging dataset, and is tolerant to the moderate rotation although it is not designed for the rotated faces. We also evaluate the speed of the three face detectors on MIDV-5OO dataset as shown in Table~\ref{tab:evalMIDV500}. The evaluation of speed is employed on CPU (Intel (R) Core(TM) i7-6500 CPU @ 2.50GHZ 2.59GHZ) and GPU (Nvidia-Titan X) respectively. Meanwhile the recall rate at 20, 50, 100, 200, 500 false positives on MIDV-500 are also shown in Table~\ref{tab:evalMIDV500}.  ~\figurename~\ref{fig:detectionexamples} shows the examples of the detection results obtained by the PCN, MTCNN and Cascade-CNN respectively. From the detected results, we can see that the MTCNN gains the best detection results in overall. It shows that the condition of detection is quite challenging, where it varies from the background of the environment, the complex illumination (such as the reflect light) and the distortion of the images introduced by the pose variation in the 3D space. Comparing to the other two methods, the Cascade-CNN has most mistakes like much more false positive detection and bad location of the coordinates of the bounding boxes. The PCN is much more better than Cascade-CNN, but in terms of the accuracy of locating the bounding boxes, it is inferior to MTCNN. It has to say, the examples does not include the extreme rotation case of the images, so for the up-right images, the MTCNN shows better performance than PCN. Comparing to the low resolution of the images in the Multi-Oriented FDDB or WiderFace datasets (from 40x40 to 400x400 maximum), the high resolution (1920x1080) of MIDV-500 images is the main reason for the low speed results. 

\begin{table*}
\caption{\label{tab:evalMIDV500} Speed and accuracy comparison between the three different methods. The MIDV-500 recall rate (\%) at 20, 50, 100, 200, 500 false positives.}
\begin{center}
\small
\begin{tabular}{|c|ccccc|cc|}
\hline
\multirow{2}{*}{Method} &
\multicolumn{5}{c|}{Recall rate at FP on MIDV-500} &
\multicolumn{2}{c|}{Speed/FPS}\\
\cline{2-8}
  & 20 & 50 & 100 & 200 & 500 & CPU & GPU  \\

\hline
PCN & 0.0027 & 0.0106 & 0.0301 & 0.0744 & 0.2533 & 1.30 & 1.236\\
MTCNN & 0.0213 & 0.1550 & 0.2383 & - & - & 1.344 & 2.71\\
CascadeCNN & 0.0053 & 0.0062 & 0.0089 & 0.0133 & 0.0221& 2.518 & 8.62\\
\hline
\end{tabular}
\end{center}
\end{table*}

\subsection{\textbf{Discussion}}
\begin{figure}[t]
\begin{center}
   \includegraphics[width=1.0\linewidth]{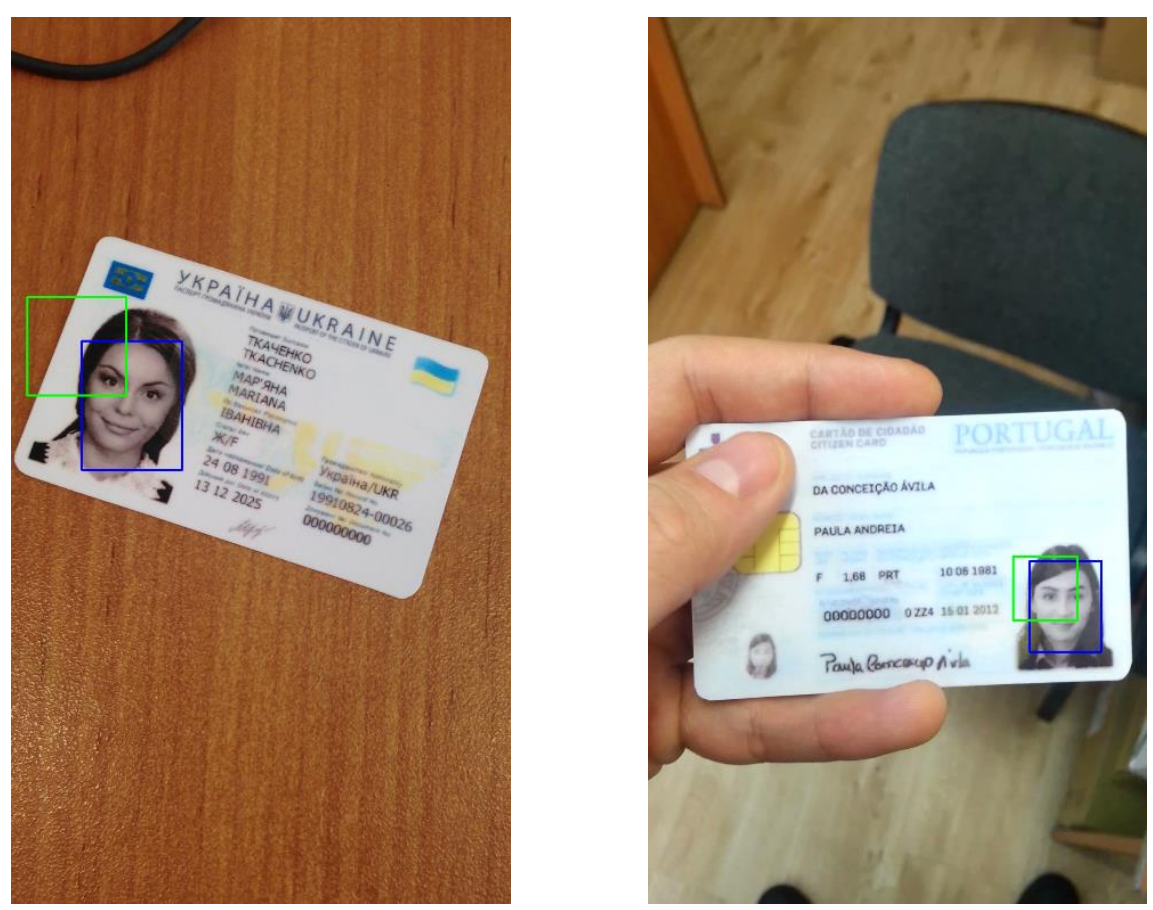}
\end{center}
   \caption{The examples show how the annotation divergence between different datasets affects the detected results. The detected bounding boxes (in green) have high confidence (larger than 0.99) but the IoU with the ground truth bounding box (in blue) are less than 0.5.}
\label{fig:annotationdiff}
\end{figure}

Comparing to the performance of these three methods on the classic datasets Multi-Oriented FDDB or WiderFace, either the recall rate or the speed of the three methods PCN, MTCNN and Casecade-CNN on the MIDV-500 dataset are greatly deteriorated. This is partly due to the quite challenging conditions of the MIDV-500 dataset, and also the heterogeneous face recognition problems. Since the methods evaluated in this work are mainly trained on the datasets such as FDDB or WiderFace which are mainly visual photos or selfies, faces in MIDV-500 dataset are the images in the ID-cards, passports, driving licences and other identity documents. These ones are way different from the training data images used in other datasets. In our case, we may find black and white photos and some relative poor image quality. However, the generalization ability is still a problem in a deep learning based data-driven detection method, which is strongly based on the data used to train the model. In particularly, for the detection task, the different way or manner used to label faces can also affect the performance. For instance, some datasets would include more parts of the head to completely cover the face, so the resulting detected bounding boxes are larger than others, this will cause the difference when we calculate the intersection-over-union used to evaluate the model. ~\figurename~\ref{fig:annotationdiff} shows how the difference of annotating bounding boxes between different datasets affects the detection. The green bounding boxes are the detected results by PCN with very high confidence (larger than 0.99), they indicate that the model \textit{thought} has detected faces very well. However, we can see that the detected bounding boxes do not really fit the ground truth (the IoU with the ground truth bounding box is less than 0.5). This divergence is probably caused by the different way to annotate the bounding boxes between MIDV-500 and the training dataset.

\section{\textbf{Conclusion}}
In this paper, we evaluated three state-of-the-art face detection methods PCN, MTCNN and Cascade-CNN on the new challenging MIDV-500 dataset. Since the face detection is a necessary step for the following identity document analysis and recognition, it is worth to survey how the current face detection methods work on the ID card-like documents such as passports, citizen cards and ID-cards. The evaluation results show the performance and the limitations of the current methods for face detection on the new challenging MIDV-500 dataset. Yet, there is still much space to improve for future works for the face detection task under challenging conditions. 

\section*{Acknowledgment}
This work was supported by the MOBIDEM project, part of the ``Systematic Paris-Region'' and ``Images \& Network'' Clusters, funded by the French government.

{\small
\newcommand{\BIBdecl}{\setlength{\itemsep}{0.25 em}}
\bibliographystyle{IEEEtran}
\bibliography{Papier_BAKKALI_CBDAR2019.bib}

\begin{thebibliography}{10}
\providecommand{\url}[1]{#1}
\csname url@samestyle\endcsname
\providecommand{\newblock}{\relax}
\providecommand{\bibinfo}[2]{#2}
\providecommand{\BIBentrySTDinterwordspacing}{\spaceskip=0pt\relax}
\providecommand{\BIBentryALTinterwordstretchfactor}{4}
\providecommand{\BIBentryALTinterwordspacing}{\spaceskip=\fontdimen2\font plus
\BIBentryALTinterwordstretchfactor\fontdimen3\font minus
  \fontdimen4\font\relax}
\providecommand{\BIBforeignlanguage}[2]{{%
\expandafter\ifx\csname l@#1\endcsname\relax
\typeout{** WARNING: IEEEtran.bst: No hyphenation pattern has been}%
\typeout{** loaded for the language `#1'. Using the pattern for}%
\typeout{** the default language instead.}%
\else
\language=\csname l@#1\endcsname
\fi
#2}}
\providecommand{\BIBdecl}{\relax}
\BIBdecl

\bibitem{de2016money}
L.~De~Koker, ``Money laundering compliance—the challenges of technology,'' in
  \emph{Financial Crimes: Psychological, Technological, and Ethical
  Issues}.\hskip 1em plus 0.5em minus 0.4em\relax Springer, 2016, pp. 329--347.

\bibitem{arlazarov2018midv}
V.~V. Arlazarov, K.~Bulatov, T.~Chernov, and V.~L. Arlazarov, ``Midv-500: A
  dataset for identity documents analysis and recognition on mobile devices in
  video stream,'' \emph{arXiv preprint arXiv:1807.05786}, 2018.

\bibitem{skoryukina2015real}
N.~Skoryukina, D.~P. Nikolaev, A.~Sheshkus, and D.~Polevoy, ``Real time
  rectangular document detection on mobile devices,'' in \emph{Seventh
  International Conference on Machine Vision (ICMV 2014)}, vol. 9445.\hskip 1em
  plus 0.5em minus 0.4em\relax International Society for Optics and Photonics,
  2015, p. 94452A.

\bibitem{usilin2010visual}
S.~Usilin, D.~Nikolaev, V.~Postnikov, and G.~Schaefer, ``Visual appearance
  based document image classification,'' in \emph{2010 IEEE International
  Conference on Image Processing}.\hskip 1em plus 0.5em minus 0.4em\relax IEEE,
  2010, pp. 2133--2136.

\bibitem{hu2017finding}
P.~Hu and D.~Ramanan, ``Finding tiny faces,'' in \emph{Proceedings of the IEEE
  conference on computer vision and pattern recognition}, 2017, pp. 951--959.

\bibitem{najibi2017ssh}
M.~Najibi, P.~Samangouei, R.~Chellappa, and L.~S. Davis, ``Ssh: Single stage
  headless face detector,'' in \emph{Proceedings of the IEEE International
  Conference on Computer Vision}, 2017, pp. 4875--4884.

\bibitem{yang2015facial}
S.~Yang, P.~Luo, C.-C. Loy, and X.~Tang, ``From facial parts responses to face
  detection: A deep learning approach,'' in \emph{Proceedings of the IEEE
  International Conference on Computer Vision}, 2015, pp. 3676--3684.

\bibitem{shi2018real}
X.~Shi, S.~Shan, M.~Kan, S.~Wu, and X.~Chen, ``Real-time rotation-invariant
  face detection with progressive calibration networks,'' in \emph{Proceedings
  of the IEEE Conference on Computer Vision and Pattern Recognition}, 2018, pp.
  2295--2303.

\bibitem{zhang2016joint}
K.~Zhang, Z.~Zhang, Z.~Li, and Y.~Qiao, ``Joint face detection and alignment
  using multitask cascaded convolutional networks,'' \emph{IEEE Signal
  Processing Letters}, vol.~23, no.~10, pp. 1499--1503, 2016.

\bibitem{li2015convolutional}
H.~Li, Z.~Lin, X.~Shen, J.~Brandt, and G.~Hua, ``A convolutional neural network
  cascade for face detection,'' in \emph{Proceedings of the IEEE conference on
  computer vision and pattern recognition}, 2015, pp. 5325--5334.

\bibitem{awal2017complex}
A.~M. Awal, N.~Ghanmi, R.~Sicre, and T.~Furon, ``Complex document
  classification and localization application on identity document images,'' in
  \emph{2017 14th IAPR International Conference on Document Analysis and
  Recognition (ICDAR)}, vol.~1.\hskip 1em plus 0.5em minus 0.4em\relax IEEE,
  2017, pp. 426--431.

\bibitem{simon2015fine}
M.~Simon, E.~Rodner, and J.~Denzler, ``Fine-grained classification of identity
  document types with only one example,'' in \emph{2015 14th IAPR International
  Conference on Machine Vision Applications (MVA)}.\hskip 1em plus 0.5em minus
  0.4em\relax IEEE, 2015, pp. 126--129.

\bibitem{de2015use}
L.-P. De~las Heras, O.~R. Terrades, J.~Llados, D.~Fernandez-Mota, and
  C.~Canero, ``Use case visual bag-of-words techniques for camera based
  identity document classification,'' in \emph{2015 13th International
  Conference on Document Analysis and Recognition (ICDAR)}.\hskip 1em plus
  0.5em minus 0.4em\relax IEEE, 2015, pp. 721--725.

\bibitem{viola2001rapid}
P.~Viola, M.~Jones \emph{et~al.}, ``Rapid object detection using a boosted
  cascade of simple features,'' \emph{CVPR (1)}, vol.~1, no. 511-518, p.~3,
  2001.

\bibitem{cinbis2011unsupervised}
R.~G. Cinbis, J.~Verbeek, and C.~Schmid, ``Unsupervised metric learning for
  face identification in tv video,'' in \emph{2011 International Conference on
  Computer Vision}.\hskip 1em plus 0.5em minus 0.4em\relax IEEE, 2011, pp.
  1559--1566.

\bibitem{wolf2011face}
L.~Wolf, T.~Hassner, and I.~Maoz, \emph{Face recognition in unconstrained
  videos with matched background similarity}.\hskip 1em plus 0.5em minus
  0.4em\relax IEEE, 2011.

\bibitem{sivic2005person}
J.~Sivic, M.~Everingham, and A.~Zisserman, ``Person spotting: video shot
  retrieval for face sets,'' in \emph{International conference on image and
  video retrieval}.\hskip 1em plus 0.5em minus 0.4em\relax Springer, 2005, pp.
  226--236.

\bibitem{lu2015surpassing}
C.~Lu and X.~Tang, ``Surpassing human-level face verification performance on
  lfw with gaussianface,'' in \emph{Twenty-ninth AAAI conference on artificial
  intelligence}, 2015.

\bibitem{starovoitov2002three}
V.~Starovoitov, D.~Samal, and D.~Briliuk, ``Three approaches for face
  recognition,'' in \emph{The 6-th International Conference on Pattern
  Recognition and Image Analysis October}, 2002, pp. 21--26.

\bibitem{starovoitov2000matching}
V.~Starovoitov, D.~Samal, and B.~Sankur, ``Matching of faces in camera images
  and document photographs,'' in \emph{2000 IEEE International Conference on
  Acoustics, Speech, and Signal Processing. Proceedings (Cat. No. 00CH37100)},
  vol.~4.\hskip 1em plus 0.5em minus 0.4em\relax IEEE, 2000, pp. 2349--2352.

\bibitem{shi2018docface}
Y.~Shi and A.~K. Jain, ``Docface: Matching id document photos to selfies,'' in
  \emph{2018 IEEE 9th International Conference on Biometrics Theory,
  Applications and Systems (BTAS)}.\hskip 1em plus 0.5em minus 0.4em\relax
  IEEE, 2018, pp. 1--8.

\bibitem{king2009dlib}
D.~E. King, ``Dlib-ml: A machine learning toolkit,'' \emph{Journal of Machine
  Learning Research}, vol.~10, no. Jul, pp. 1755--1758, 2009.

\bibitem{bradski2000opencv}
G.~Bradski, ``The opencv library. dr. dobb's j. of softw. tools,'' 2000.

\bibitem{jain2010fddb}
V.~Jain and E.~Learned-Miller, ``Fddb: A benchmark for face detection in
  unconstrained settings,'' 2010.

\bibitem{yang2016wider}
S.~Yang, P.~Luo, C.-C. Loy, and X.~Tang, ``Wider face: A face detection
  benchmark,'' in \emph{Proceedings of the IEEE conference on computer vision
  and pattern recognition}, 2016, pp. 5525--5533.

\bibitem{zhu2012face}
X.~Zhu and D.~Ramanan, ``Face detection, pose estimation, and landmark
  localization in the wild,'' in \emph{2012 IEEE conference on computer vision
  and pattern recognition}.\hskip 1em plus 0.5em minus 0.4em\relax IEEE, 2012,
  pp. 2879--2886.

\bibitem{opitz2016grid}
M.~Opitz, G.~Waltner, G.~Poier, H.~Possegger, and H.~Bischof, ``Grid loss:
  Detecting occluded faces,'' in \emph{European conference on computer
  vision}.\hskip 1em plus 0.5em minus 0.4em\relax Springer, 2016, pp. 386--402.

\end{thebibliography}
}

\end{document}